\documentclass[11pt, letterpaper]{article}

\usepackage{amsmath}
\usepackage{amssymb}
\usepackage{amsthm}
\usepackage{graphicx}
\usepackage{hyperref}
\usepackage{url}
\usepackage{algorithm}
\usepackage{algorithmic}
\usepackage{caption}
\usepackage{subcaption}
\usepackage{booktabs}
\usepackage{authblk}
\usepackage{appendix}
\usepackage[utf8]{inputenc}
\usepackage[T1]{fontenc}
\usepackage[english]{babel}
\usepackage{geometry}
\usepackage{xcolor}
\usepackage{fancyhdr}
\usepackage{microtype}
\usepackage{parskip}
\usepackage{tabularx}
\usepackage{enumitem}

\hypersetup{
    colorlinks=true,
    linkcolor=blue,
    filecolor=magenta,      
    urlcolor=cyan,
    citecolor=red,
}

\definecolor{headercolor}{RGB}{0, 50, 100}
\title{Differentiate ChatGPT-generated and Human-written Medical Texts}

\author[1]{Wenxiong Liao}
\author[2]{Zhengliang Liu}
\author[2]{Haixing Dai}
\author[2]{Shaochen Xu}
\author[2]{Zihao Wu}
\author[1]{Yiyang Zhang}
\author[1]{Xiaoke Huang}
\author[3]{Dajiang Zhu}
\author[1]{Hongmin Cai}
\author[2]{Tianming Liu}
\author[4]{Xiang Li}

\affil[1]{School of Computer Science and Engineering, South China University of Technology}
\affil[2]{School of Computing, University of Georgia}
\affil[3]{University of Texas at Arlington}
\affil[4]{Massachusetts General Hospital and Harvard Medical School}

\date{}

\begin{document}

\maketitle

\begin{abstract}

\textbf{Background:} Large language models such as ChatGPT are capable of generating grammatically perfect and human-like text content, and a large number of ChatGPT-generated texts have appeared on the Internet. However, medical texts such as clinical notes and diagnoses require rigorous validation, and erroneous medical content generated by ChatGPT could potentially lead to disinformation that poses significant harm to healthcare and the general public. 

\textbf{Objective:} This research is among the first studies on responsible and ethical AIGC (Artificial Intelligence Generated Content) in medicine. We focus on analyzing the differences between medical texts written by human experts and generated by ChatGPT, and designing machine learning workflows to effectively detect and differentiate medical texts generated by ChatGPT.

\textbf{Methods:} We first construct a suite of datasets containing medical texts written by human experts and generated by ChatGPT. In the next step, we analyze the linguistic features of these two types of content and uncover differences in vocabulary, part-of-speech, dependency, sentiment, perplexity, etc. Finally, we design and implement machine learning methods to detect medical text generated by ChatGPT.

\textbf{Results:} Medical texts written by humans are more concrete, more diverse, and typically contain more useful information, while medical texts generated by ChatGPT pay more attention to fluency and logic, and usually express general terminologies rather than effective information specific to the context of the problem. A BERT-based model can effectively detect medical texts generated by ChatGPT, and the F1 exceeds 95\%.

\textbf{Conclusions:} Although text generated by ChatGPT is grammatically perfect and human-like, the linguistic characteristics of generated medical texts are different from those written by human experts. Medical text generated by ChatGPT can be effectively detected by the proposed machine learning algorithms. This study provides a pathway towards trustworthy and accountable use of large language models in medicine.


\end{abstract}

\section{Introduction}
\subsection{Background}
Since the advent of pre-trained language models such as GPT (Generative Pre-trained Transformer) \cite{radford2018improving} and BERT (Bidirectional Encoder Representations from Transformers) \cite{kenton2019bert}  in 2018, transformer-based \cite{vaswani2017attention} language models have revolutionized and popularized NLP. More recently, (very) large language models (LLM) \cite{brown2020language,ouyang2022training} have demonstrated superior performance on zero-shot and few-shot tasks. Among large language models, ChatGPT is favored by users due to its accessibility, as well as its ability to produce grammatically correct and human-level answers in different domains. Since the release of ChatGPT in November 2022 by OpenAI, it has quickly gained significant attention within a few months and has been widely discussed in the natural language processing (NLP) community and other fields.

To balance the cost and efficiency of data annotation, and train a large language model which that better aligns with user intent in a helpful and safe manner, researchers used reinforcement learning from human feedback (RLHF) \cite{christiano2017deep}  to develop ChatGPT. The RLHF uses a ranking-based human preference dataset to train a reward model and with this reward model, ChatGPT can be fine-tuned by proximal policy optimization (PPO) \cite{schulman2017proximal}. As a result, ChatGPT can understand the meaning and intent behind user queries, which empowers ChatGPT to respond to queries in the most relevant and useful way. In addition to aligning with user intent, another factor that makes ChatGPT popular is its ability to handle a variety of tasks in different domains. The massive training corpus from the world wide web endows ChatGPT with the ability to learn the nuances of human language patterns. ChatGPT seems to be able to successfully generate human-level text content in all domains \cite{susnjak2023applying, dai2023chataug, wei2023zero, liu2023deid}.

However, ChatGPT is a double-edged sword \cite{shen2023chatgpt,hisan2023chatgpt}. Misusing ChatGPT to generate human-like content can easily mislead users, resulting in wrong and potentially detrimental decisions. For example, malicious actors can use ChatGPT to generate a large number of fake reviews that damage the reputation of high-quality restaurants while falsely boosting the reputation of low-quality competitors. This is an example that can potentially harm consumers \cite{mitrovic2023chatgpt}. 

ChatGPT has also demonstrated a strong understanding of high-stake domains such as medicine \cite{gilson2023does}, including specialties such as radiation oncology. \cite{holmes2023evaluating}.  Medical information typically requires rigorous validation. Indeed, false medical-related information generated by ChatGPT can easily lead to misjudgment of the developmental trend of diseases, delay the treatment process, or negatively affect the life and health of patients \cite{bickmore2018patient}.

\subsection{Development of Language Models }

The transformer-based language models have demonstrated a strong language modeling ability. Generally speaking, transformer-based language models are divided into 3 categories: encoder-based models (e.g., BERT \cite{kenton2019bert}, Roberta \cite{liu2019roberta}, Albert \cite{lan2019albert} ), decoder-based models (eg: GPT \cite{radford2018improving}, GPT2 \cite{radford2019language}), encoder-decoder-based models (e.g. Transformers \cite{vaswani2017attention}, BART \cite{lewis2020bart}, T5 \cite{raffel2020exploring}). In order to combine biomedical knowledge with language models, many researchers have added biomedical corpus for training \cite{liao2023mask,cai2022coarse,liu2022survey,liu2023summary,zhao2023brain}. Alsentzer et al. \cite{alsentzer2019publicly} fine-tuned the publicly release BERT model on the MIMIC dataset \cite{johnson2016mimic}, and demonstrated good performance on natural language inference and named entity recognition tasks. Lee et al. \cite{lee2020biobert} fine-tuned BERT on PubMed dataset and perform well on biomedical named entity recognition,  biomedical relation extraction, and biomedical question-answering tasks. Based on the backbone of GPT2 \cite{radford2019language}, Luo et al. \cite{luo2022biogpt} continue pre-training on the bio-medical dataset and show superior performance on six biomedical NLP tasks. Other innovative applications include AgriBERT \cite{rezayi2022agribert} for agriculture, ClinicalRadioBERT for radiation oncology \cite{rezayi2022clinicalradiobert} and SciEdBERT for science education \cite{liu2023context}. 

In recent years, decoder-based LLM has demonstrated excellent performance on a variety of tasks \cite{holmes2023evaluating,liu2023deid,dai2023chataug}. Compared with previous language models, LLM contains a large number of trainable parameters, such as GPT3 contains 175 billion parameters. The increased model size of GPT-3 makes it more powerful than previous models; boosting its language ability to near human levels \cite{sezgin2022operationalizing}.  The ChatGPT belongs to the GPT-3.5 series, which fine-tuned its base on RLHF. Research \cite{gilson2023does} shows that ChatGPT achieves a passing score equivalent to that of a third-year medical student on a medical question-answering task.

\subsection{Potential risks of using ChatGPT}

There is more and more content generated by ChatGPT on the Internet. However, when using ChatGPT, some potential risks need to be considered. First of all, ChatGPT may limit human creativity. ChatGPT has the ability to debug code or write essays for college students. It is important to consider whether ChatGPT will generate unique creative work, or simply copy content from their training set. New York City public schools have banned ChatGPT.

Secondly, ChatGPT with the ability to produce a text of surprising quality which can deceive readers, and the end result is a dangerous accumulation of misinformation \cite{homolak2023opportunities}. StackOverflow, a popular platform for coders and programmers, banned the use of ChatGPT-generated content. Because the average rate of correct answers from ChatGPT is too low and could cause significant harm to the site and the users who rely on it for accurate answers.

Thirdly, ChatGPT lacks the knowledge and expertise necessary to accurately and adequately convey complex scientific concepts and information. For example, human medical writers cannot yet be fully replaced because ChatGPT do not have the same level of understanding and expertise in the medical field \cite{biswas2023chatgpt}. Additionally, human medical writers will be responsible for ensuring the accuracy and completeness of the information communicated and for complying with ethical and regulatory guidelines, however, ChatGPT cannot be held responsible.

In addition, the application of ChatGPT to generate medical texts must consider some ethical issues. First of all, training a large language model requires a huge amount of data, but the high quality of the data is difficult to guarantee, so the trained ChatGPT is biased. For example, ChatGPT can provide biased output and perpetuate sexist stereotypes \cite{patel2023chatgpt}. Secondly, ChatGPT may lead to private information leakage. This may be because the large language model remembers the personal privacy information in the training set \cite{carlini2021extracting}. Thirdly, it involves the legal framework. Who is to be held accountable when an AI doctor makes an inevitable mistake? ChatGPT cannot be held accountable for its work, and there is no legal framework to determine who owns the rights to AI-generated work \cite{homolak2023opportunities}.

To prevent the misuse use of ChatGPT to generate medical texts and avoid the potential ethical risks of using ChatGPT, in this paper, we focus on the detection of ChatGPT-generated text for the medical domain. We collect both publicly-available expert-generated medical content and ChatGPT-generated content through OpenAI API. The aim of this study is twofold: 1) What is the difference between medical content written by humans and generated by ChatGPT? Can we use machine learning methods to detect whether medical content is written by human experts or ChatGPT?

In this work, we  make the following contributions to academia and industry:

\begin{itemize}
	\item We construct two datasets to analyze the difference between ChatGPT and human-generated medical text. We will release these two datasets to facilitate further analysis and research on ChatGPT for researchers.

    \item In this paper, we conducted a language analysis of the medical content written by humans and the medical content generated by ChatGPT. From the analysis results, we can grasp the difference between ChatGPT and humans in constructing medical content.

    \item  We built a variety of machine learning models to detect samples generated by humans and ChatGPT and explained and visualized the model structures.

\end{itemize}

In summary, this study is among the first efforts to qualitatively and quantitatively analyze and categorize differences between medical text from human experts and AIGC. We believe this work can spur further research in this direction and provide pathways toward responsible AIGC in medicine. 

\section{Methods}

\subsection{Dataset construction}

\begin{figure*}[h]
	\centering
	\includegraphics[width=0.8\textwidth]{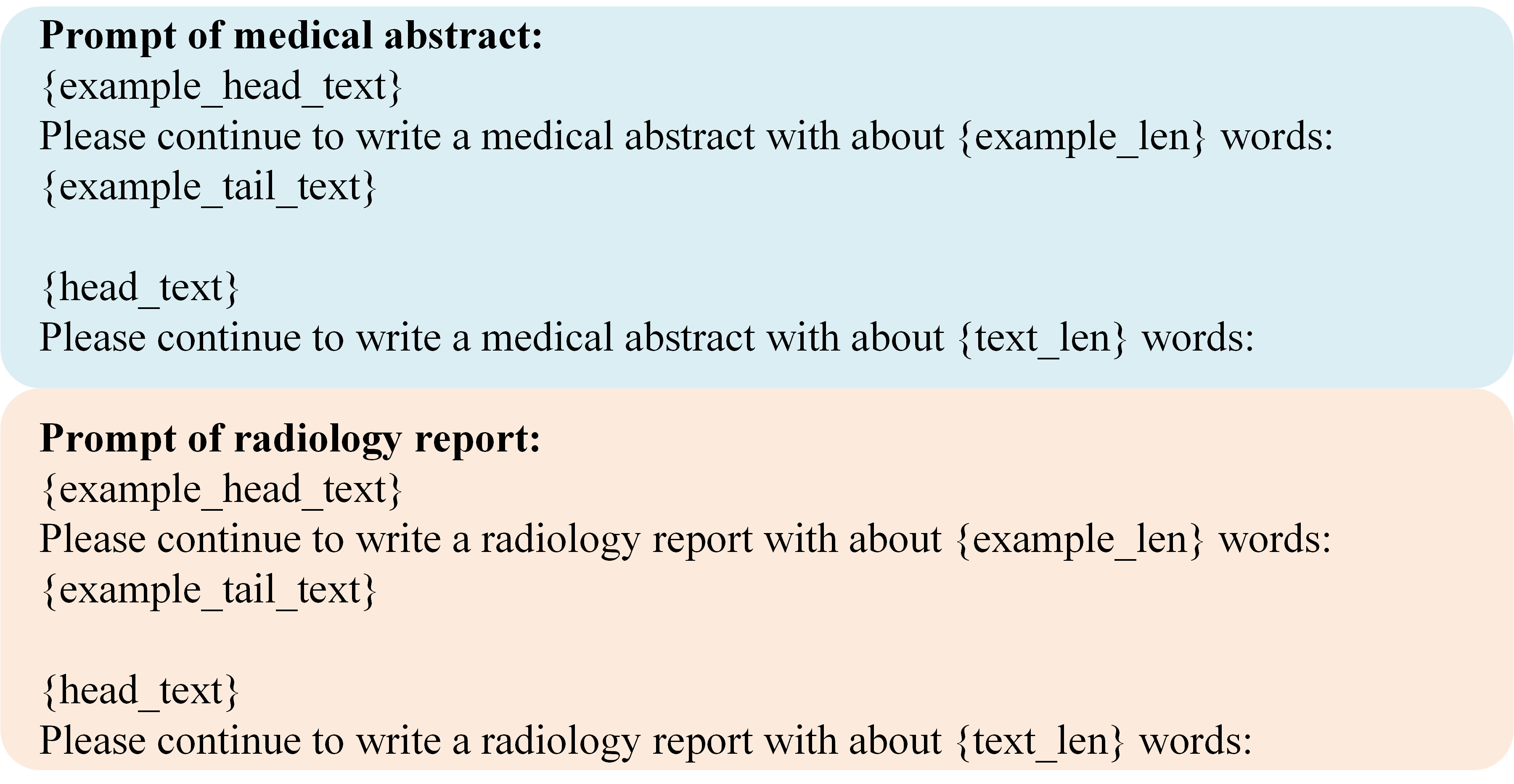} 
	\caption{The prompts of medical abstract and radiology report datasets.}
	\label{fig1}
\end{figure*}

\begin{figure*}[h]
	\centering
	\includegraphics[width=1.0\textwidth]{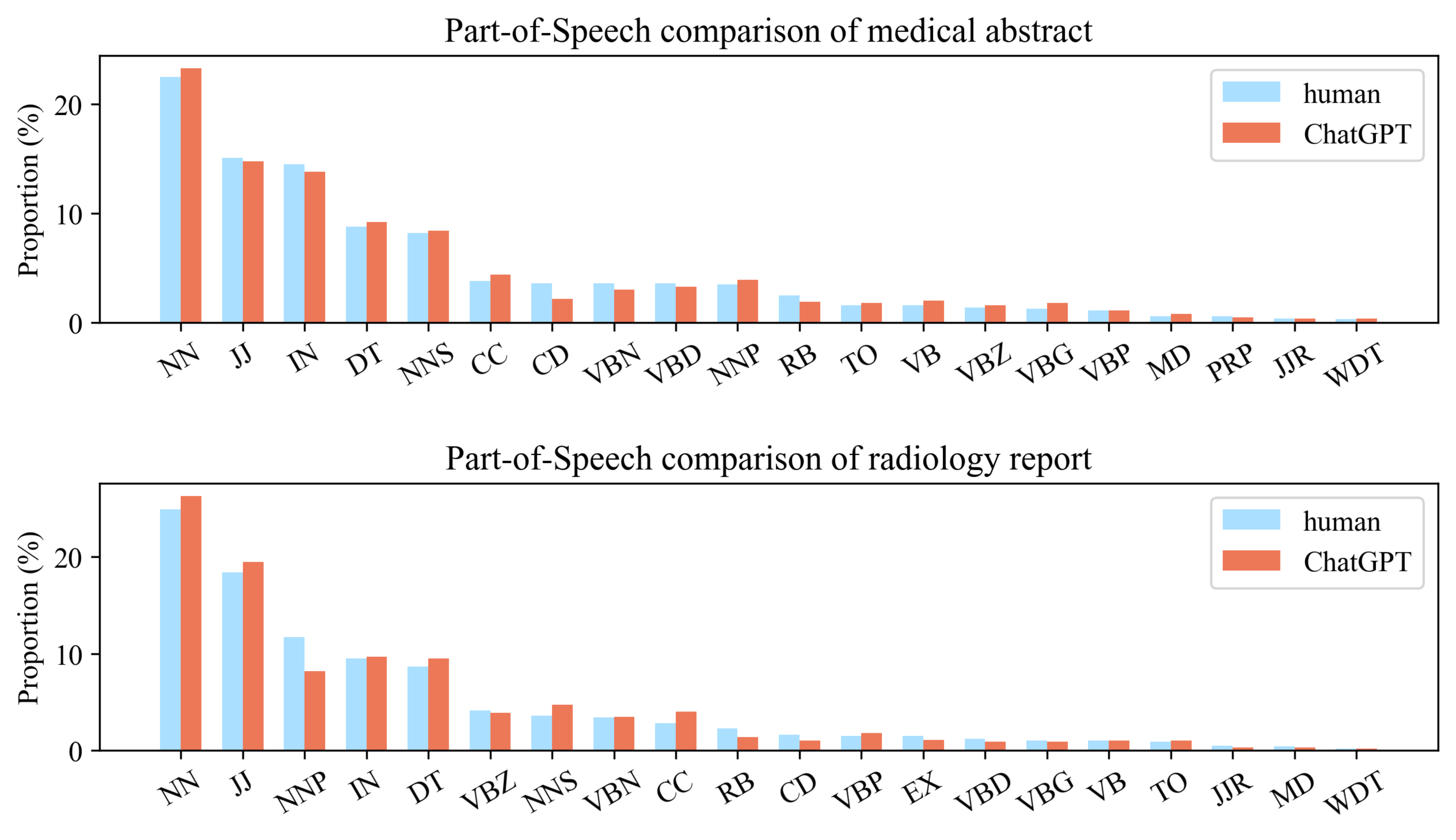} 
	\caption{part-of-speech analysis between human-written and ChatGPT generated medical text. }
	\label{fig2}
\end{figure*}

To analyze and discriminate human and ChatGPT-generated medical texts, we constructed two datasets:

\begin{itemize}
    \item \textbf{medical abstract}: This original dataset comes from kaggle\footnote{https://www.kaggle.com/datasets/chaitanyakck/medical-text}.  The medical dataset involves 5 different conditions: digestive system diseases, cardiovascular diseases, neoplasms, nervous system diseases, and general pathological conditions.

    \item \textbf{radiology report}: This original dataset comes from the work of Johnson et al. \cite{johnson2016mimic}, and we only select a part of radiology reports to build our radiology report dataset.

\end{itemize}

We sampled 2200 samples from the medical abstract and radiology report datasets as medical texts written by humans. In order to guide ChatGPT to generate medical content, we adopt the method of text continuation with demonstration instead of rephrase \cite{mitrovic2023chatgpt} or query \cite{guo2023close}, with in-context learning, because text continuation can produce more human-like text. The prompts of medical abstract and radiology report datasets are shown in Figure \ref{fig1}. We randomly select a sample (except the sample itself) from the dataset as a demonstration. Finally, we obtained medical abstract and radiology report datasets containing 4400 samples respectively.

\subsection{Linguistic analysis}

we will perform linguistic analysis of the medical content generated by humans and ChatGPT, including vocabulary and sentence feature analysis, part-of-speech (POS) analysis, dependency parsing, sentiment analysis, and text perplexity. 

The vocabulary and sentence feature analysis illuminates the differences in the statistical characteristics of the words and sentences constructed by humans and ChatGPT when generating medical texts. We use NLTK (Natural Language Toolkit) \cite{bird2009natural} to perform POS analysis. Dependency parsing is a technique that analyzes the grammatical structure of a sentence by identifying the dependencies between the words of the sentence.  We apply stanford-corenlp for dependency parsing and compare the proportions of different dependency relationships and their corresponding dependency distances. We apply a pre-trained sentiment analysis model \footnote{https://huggingface.co/cardiffnlp/twitter-roberta-base-sentiment} to conduct sentiment analysis for both medical abstract and radiology report datasets. Perplexity is often used as a metric to evaluate the performance of a language model, with lower perplexity indicating that the language model is more confident in its predictions. We use the BioGPT \cite{luo2022biogpt} model to compute the perplexity of the human-written and ChatGPT-generated medical text.

\begin{figure*}[h]
	\centering
	\includegraphics[width=1.0\textwidth]{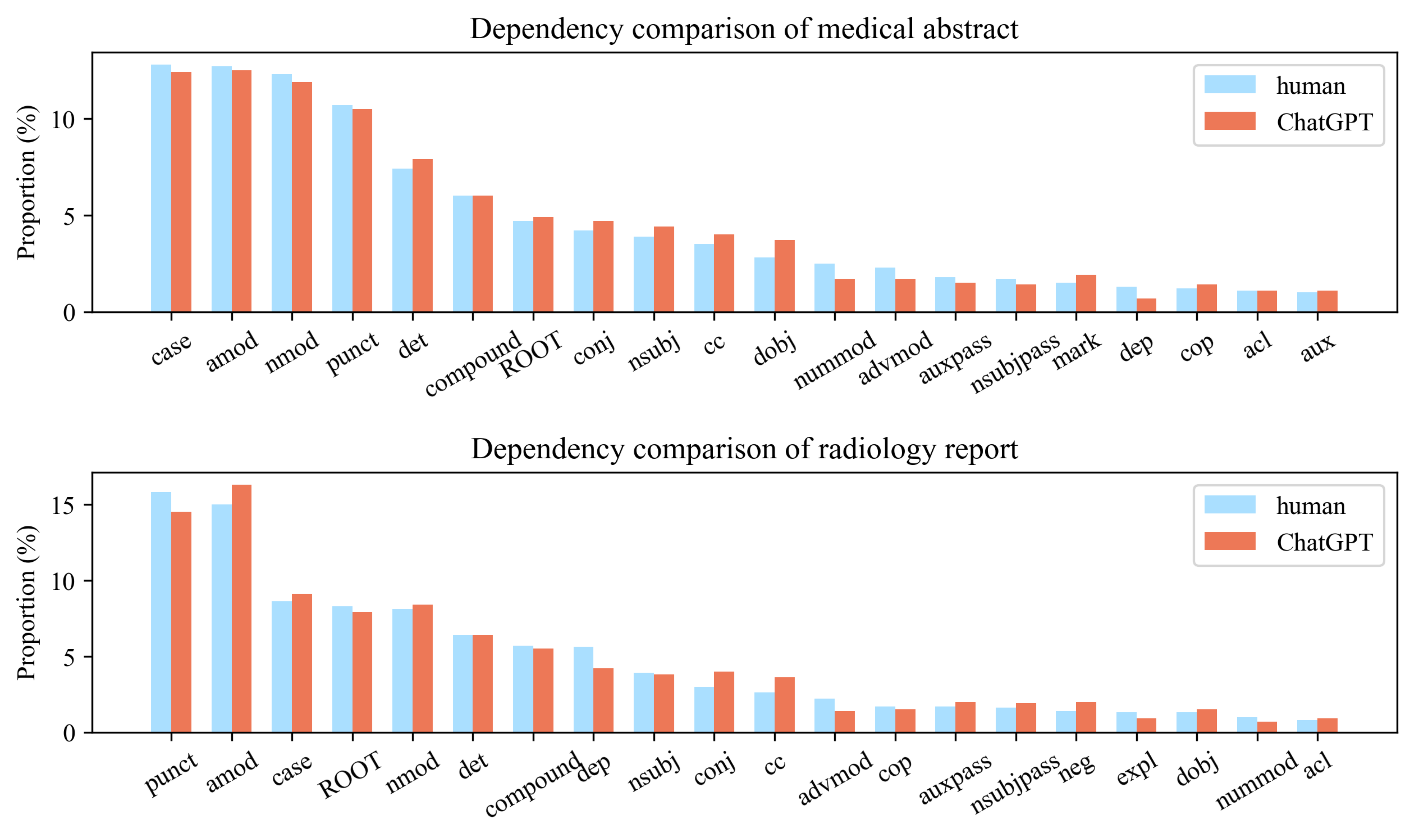} 
	\caption{ Top-20 dependency relations between human-written and ChatGPT generated medical text. }
	\label{fig3}
\end{figure*}

\begin{figure*}[h]
	\centering
	\includegraphics[width=1.0\textwidth]{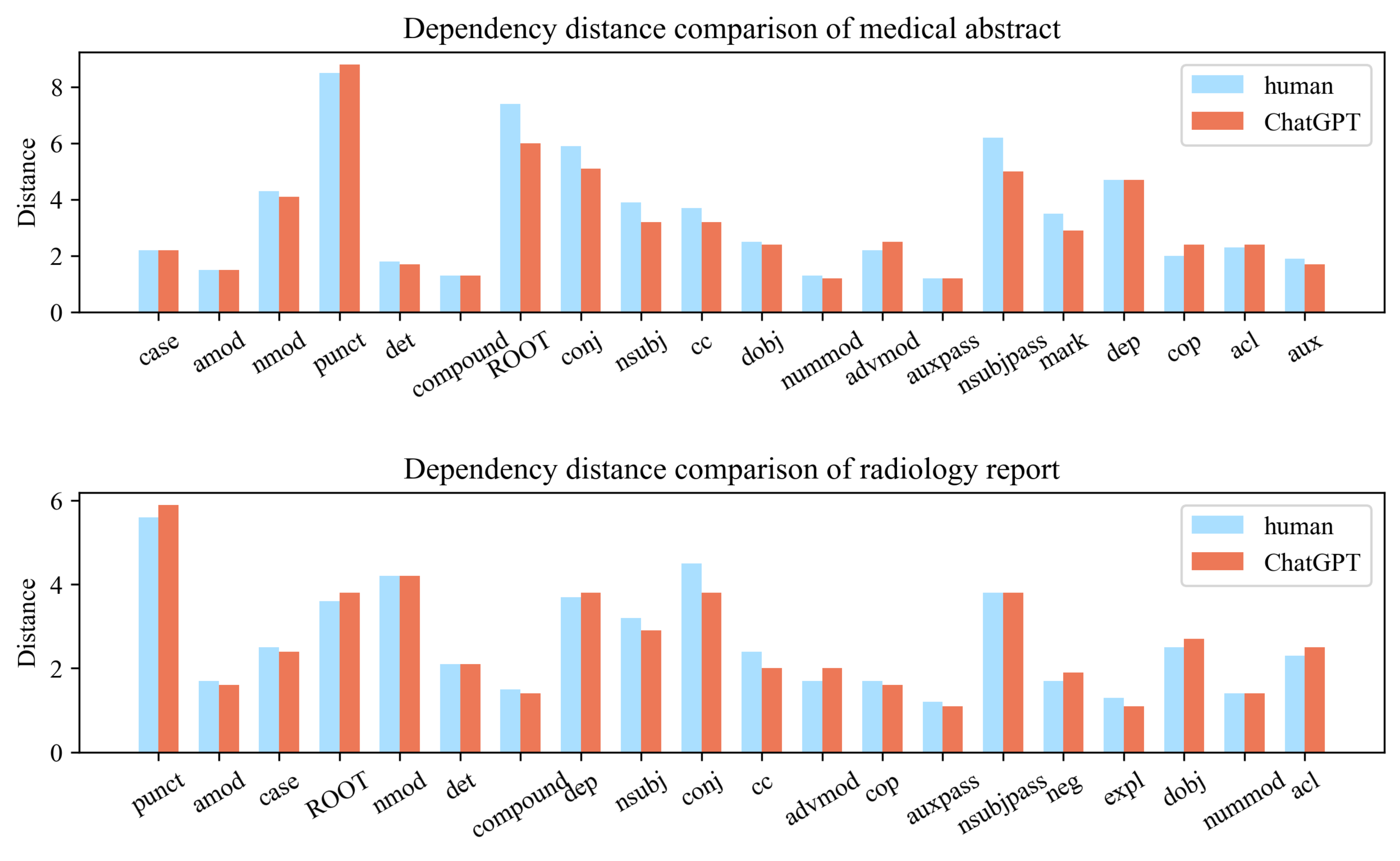} 
	\caption{Top-20 dependency distances relations between human-written and ChatGPT generated medical text. }
	\label{fig4}
\end{figure*}

\subsection{Detect ChatGPT-generated Texts}

The text content generated by the LLM has become popular on the Internet. Since most of the content generated by the LLM is text with a fixed language pattern and language style, when a large number of generated text content appears, it will not be conducive to human active creation, and It can also cause panic if the incorrect medical text is generated. We  use a variety of methods to detect medical texts generated by ChatGPT to reduce the potential risks to society caused by improper or malicious use of language models.

 First, we divide the medical abstract and radiology report datasets into a training set, test set, and validation set at a ratio of 7:2:1, respectively. Then we use a variety of algorithms to model in the training set, select the best model parameters through the validation set, and finally calculate the metrics on the test set.

 \begin{itemize}
    \item \textbf{Perplexity-CLS}: As shown in Figure 6, the medical text written by humans with higher text perplexity than the medical text generated by ChatGPT. An intuitive idea is to find an optimal perplexity threshold to detect the medical text generated by ChatGPT. The idea is the same as GPTZero \footnote{https://gptzero.me/}, but our data is medical-related text, so we use BioGPT \cite{luo2022biogpt} as a language model to calculate text perplexity. We find the optimal perplexity threshold in the validation set and calculate the metrics on the test set.

    \item \textbf{CART} (Classification and Regression Trees): CART is a classic decision tree algorithm, the classification tree uses gini as the measure of feature division. We vectorize the samples through TF-IDF (term frequency–inverse document frequency), and for the convenience of visualization, we set the maximum depth of the tree to 4.

    \item \textbf{XGBoost} \cite{chen2016xgboost}: XGBoost is an ensemble learning method, and we set the maximum  depth for base learners as 4 and vectorize the samples by TF-IDF.

    \item \textbf{BERT} \cite{kenton2019bert}: BERT is a pre-trained language model, we fine-tune our medical text based on pre-trained BERT to detect the text generated by ChatGPT. The version of BERT we use is bert-base-cased \footnote{https://huggingface.co/bert-base-cased}.
\end{itemize}

In addition, we will analyze the models of CART, XGBoost, and BERT to explore what features of the text help to detect the text generated by ChatGPT. 

\section{Results}

\subsection{Linguistic analysis}

	\begin{table}[h]
		\centering
\begin{tabular}{lllllllll}
\hline
{\color[HTML]{333333} }                 & \multicolumn{2}{c}{{\color[HTML]{333333} vocabulary}}       & \multicolumn{2}{c}{{\color[HTML]{333333} \#stem}}             & \multicolumn{2}{c}{{\color[HTML]{333333} \#sentence}}         & \multicolumn{2}{c}{{\color[HTML]{333333} sentence length}}    \\
{\color[HTML]{333333} }                 & {\color[HTML]{333333} human} & {\color[HTML]{333333} chatgpt} & {\color[HTML]{333333} human} & {\color[HTML]{333333} chatgpt} & {\color[HTML]{333333} human} & {\color[HTML]{333333} chatgpt} & {\color[HTML]{333333} human} & {\color[HTML]{333333} chatgpt} \\ \hline
{\color[HTML]{333333} medical abstract} & {\color[HTML]{333333} 25186} & {\color[HTML]{333333} 19260}   & {\color[HTML]{333333} 17788} & {\color[HTML]{333333} 13631}   & {\color[HTML]{333333} 8.7}   & {\color[HTML]{333333} 9.6}     & {\color[HTML]{333333} 16.2}  & {\color[HTML]{333333} 15.9}    \\
{\color[HTML]{333333} radiology report} & {\color[HTML]{333333} 11343} & {\color[HTML]{333333} 8976}    & {\color[HTML]{333333} 8522}  & {\color[HTML]{333333} 6631}    & {\color[HTML]{333333} 12.7}  & {\color[HTML]{333333} 11.8}    & {\color[HTML]{333333} 9.7}   & {\color[HTML]{333333} 10.4}    \\ \hline
\end{tabular}
		\caption{vocabulary and sentence  analysis. \#stem is the number of word stem, and \#sentence is the average number of sentences per sample.}
		\label{table1}
	\end{table}

\textbf{Vocabulary and sentence  analysis:} As shown in Table \ref{table1}, From the perspective of statistical characteristics, the main difference between the human written medical text and the medical text generated by ChatGPT exists in the vocabulary and stem. Human-written medical text vocabulary size and the number of stems are significantly larger than those of ChatGPT. This suggests that the content and expression of medical texts written by humans are more diverse, which is more in line with the actual patient situation, while the texts generated by ChatGPT are more inclined to use commonly used words to express common situations.

\textbf{Part-of-speech analysis:} The results of POS  are shown in Figure \ref{fig2}. The ChatGPT uses more words about  NN(noun), DT(determiner), NNS(noun plural), and CC(coordinating conjunction), while using less CD(cardinal-digit) and RB(adverb). 

Frequent use of NN and NNS tends to indicate that the text is more argumentative, showing information and objectivity \cite{nagy2012words}. The high proportion of CC and DT indicates that the structure of the medical text and the relationship between causality, progression, or contrast is clear. At the same time, a large number of CDs and RBs appear in medical texts written by humans, indicating that the expressions are more specific rather than general. For example, doctors will use specific numbers to describe the size of tumors.

\textbf{Dependency parsing:} The results of dependency parsing are shown in Figure \ref{fig3} and Figure \ref{fig4}.  As shown in Figure 3, the comparison of dependencies exhibits similar characteristics to POS analysis, where ChatGPT uses more det (determiner), conj (conjunct), and cc(coordination ) relations while using less nummod (numeric modifier) and advmod(adverbial modifier). For dependency distance, ChatGPT with obviously shorter conj(conjunct), cc(coordination), and nsubi(nominal subject) which  makes the text generated by chatGPT more logical and fluent.

\begin{figure*}[h]
	\centering
	\includegraphics[width=1.0\textwidth]{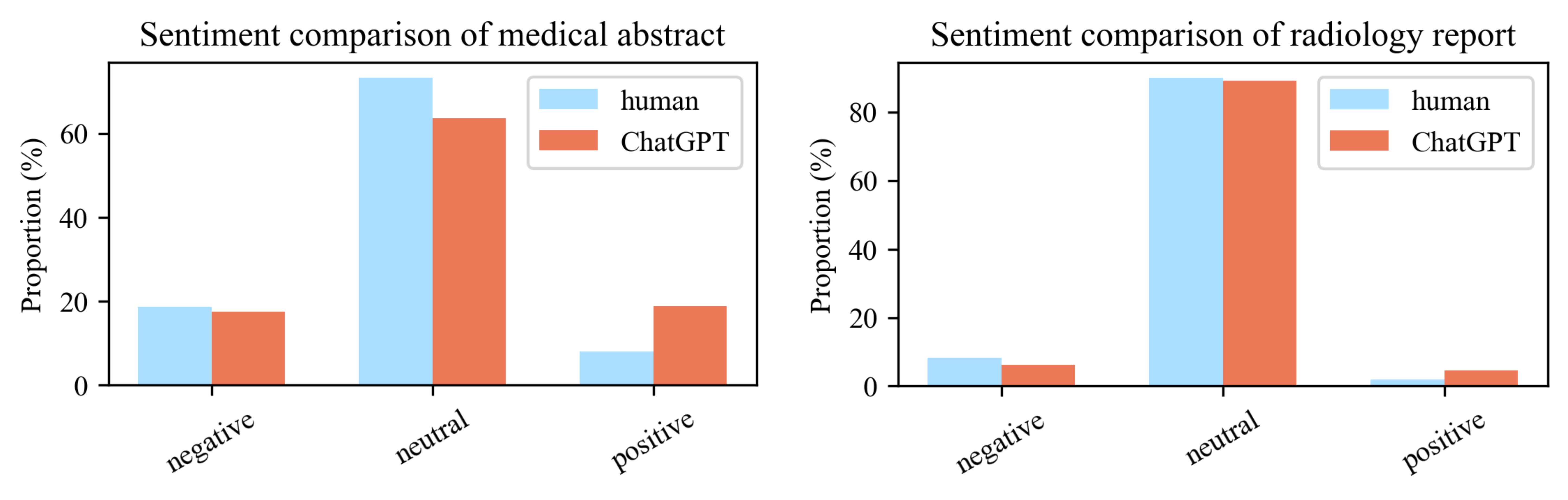} 
	\caption{Sentiment analysis of human-written and ChatGPT generated medical text. }
	\label{fig5}
\end{figure*}

\textbf{Sentiment analysis:} The results of sentiment analysis are shown in Figure \ref{fig5}. most of the medical texts written by humans or the texts generated by ChatGPT with neutral sentiments. It should be noted that the proportion of negative sentiments in humans is significantly higher than that in ChatGPT, while the proportion of positive sentiments in humans is significantly lower than that in ChatGPT. This may be because ChatGPT has added a special mechanism to carefully filter the original training dataset to ensure any violent or sexual content is removed, making the generated text more neutral or positive.

\begin{figure*}[h]
	\centering
	\includegraphics[width=1.0\textwidth]{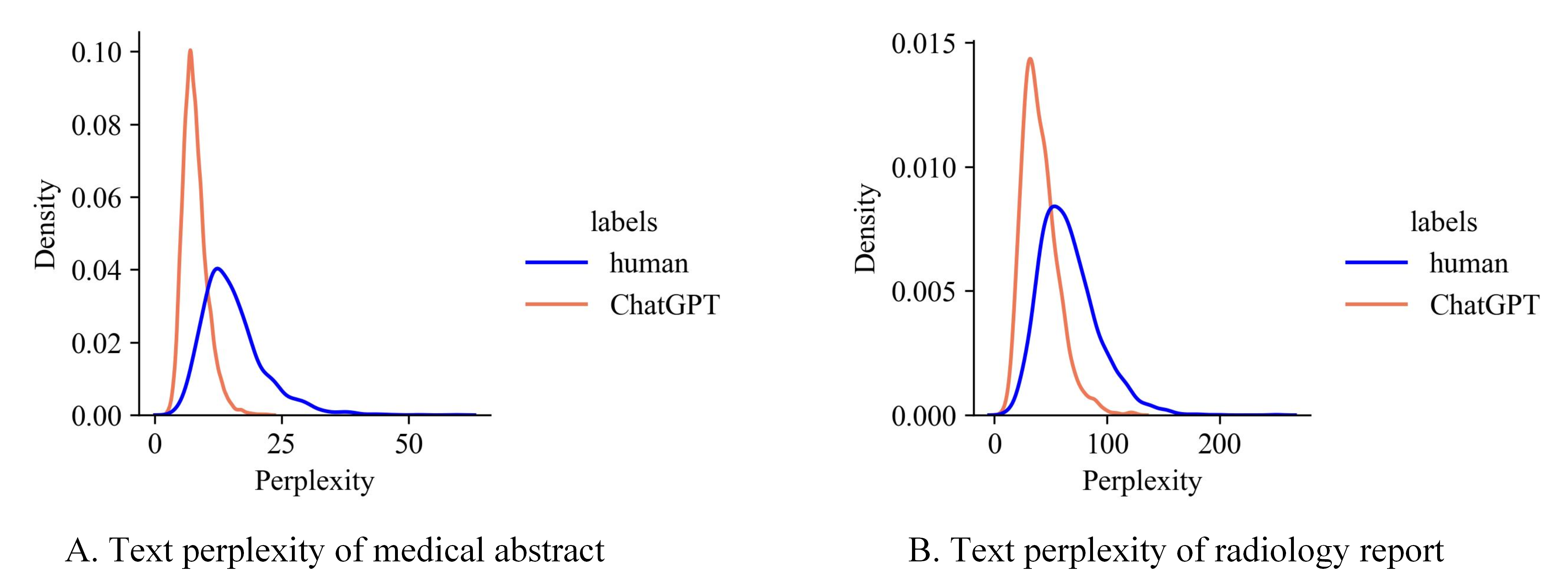} 
	\caption{Text perplexity of human-written and ChatGPT generated medical text. }
	\label{fig6}
\end{figure*}

\textbf{Text perplexity:} The results of text perplexity are shown in Figure \ref{fig6}. It can be observed that whether it is a medical abstract or a radiation report dataset, the text perplexity generated by ChatGPT is significantly lower than that written by humans. ChatGPT captures common patterns and structures in the training corpus and is very good at replicating them. Therefore, the text generated by ChatGPT has relatively low perplexity. Humans can express themselves in a variety of ways, depending on the intellectual context, the condition of the patient, etc., which may make BioGPT more difficult to predict. Therefore, human-written text with a higher perplexity and  wider distribution.

Through the above analysis, we can get the main differences between the human-written  and ChatGPT-generated  medical text, including:

\begin{itemize}
    \item Medical texts written by humans are more diverse, while medical texts generated by ChatGPT are more common.

    \item Medical texts generated by ChatGPT have better logic and fluency.

    \item Medical texts written by humans contain more specific values and text content is more specific.

    \item Medical texts generated by ChatGPT are more neutral and positive.

    \item ChatGPT has lower text perplexity because it is good at replicating common expression patterns and sentence structures.

\end{itemize}

\section{Detect ChatGPT-generated texts}

\begin{table}[h]
		\centering
            \resizebox{1.0\columnwidth}{!}{
            \begin{tabular}{lllllllllllll}
\hline
                 & \multicolumn{3}{c}{Perplexity-CLS} & \multicolumn{3}{c}{CART}   & \multicolumn{3}{c}{XGBoost} & \multicolumn{3}{c}{BERT}   \\
                 & precision    & recall    & F1      & precision & recall & F1    & precision  & recall & F1    & precision & recall & F1    \\ \hline
medical abstract & 0.728        & 0.724     & 0.723   & 0.777     & 0.745  & 0.738 & 0.898      & 0.893  & 0.893 & 0.958     & 0.958  & 0.958 \\
radiology report  & 0.831        & 0.828     & 0.828   & 0.829     & 0.825  & 0.824 & 0.899      & 0.898  & 0.898 & 0.968     & 0.967  & 0.967 \\ \hline
                \end{tabular}
            }
		\caption{Results of detecting ChatGPT-generated medical text.}
		\label{table2}
	\end{table}

 \begin{figure*}[h]
	\centering
	\includegraphics[width=1.0\textwidth]{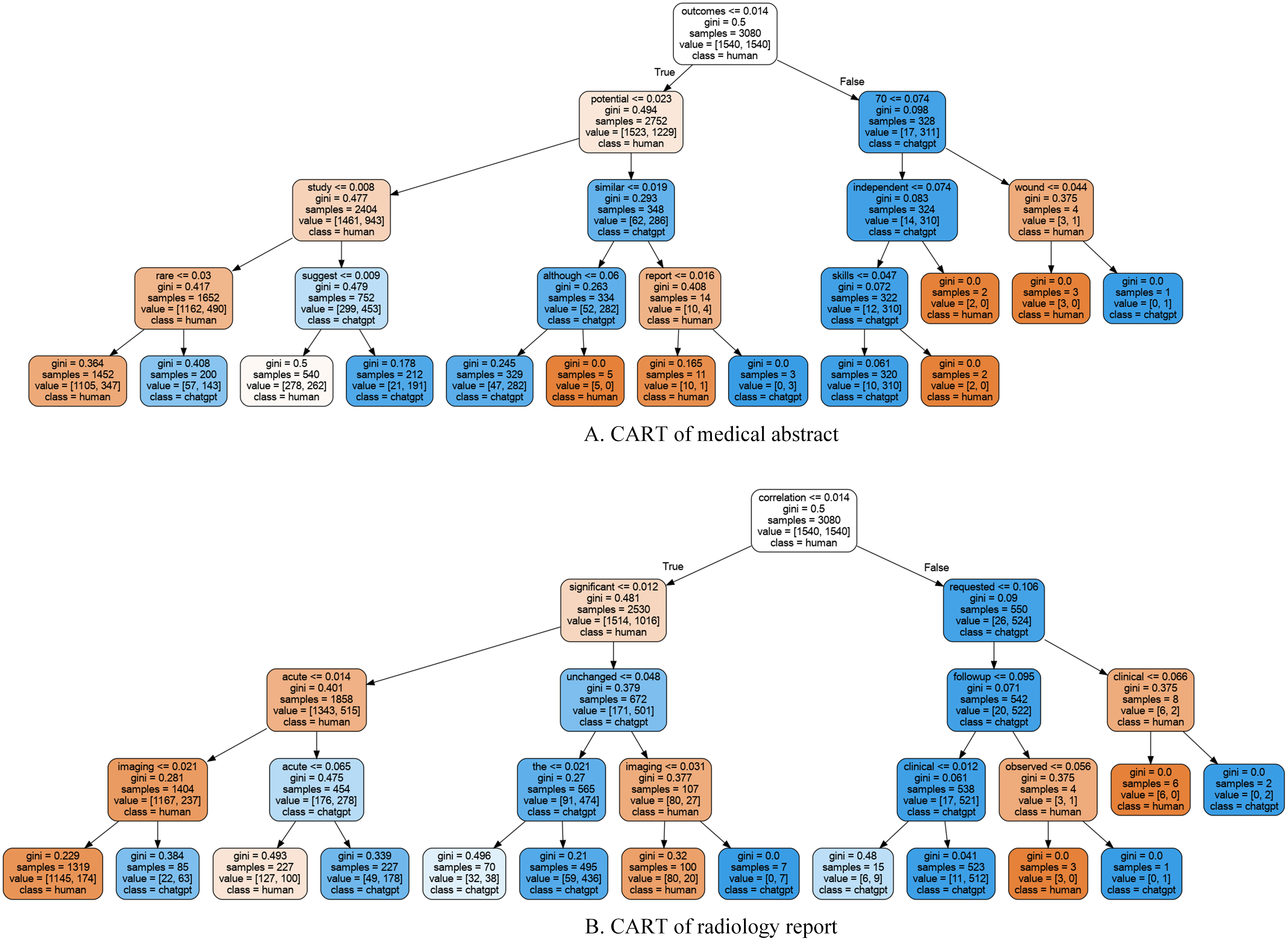} 
	\caption{Visualization of CART. }
	\label{fig7}
\end{figure*}

The results of detecting ChatGPT-generated medical text are shown in Table \ref{table2}. Since Perplexity-CLS is an unsupervised learning method, it is less effective than other methods. The XGBoost integrates the results of multiple decision trees, so it works better than CART with a single decision tree. The pre-trained BERT model can easily recognize the differences in the logical structure and language style of medical texts written by humans and generated by ChatGPT, thus achieving the best performance.

\begin{figure*}[h]
	\centering
	\includegraphics[width=1.0\textwidth]{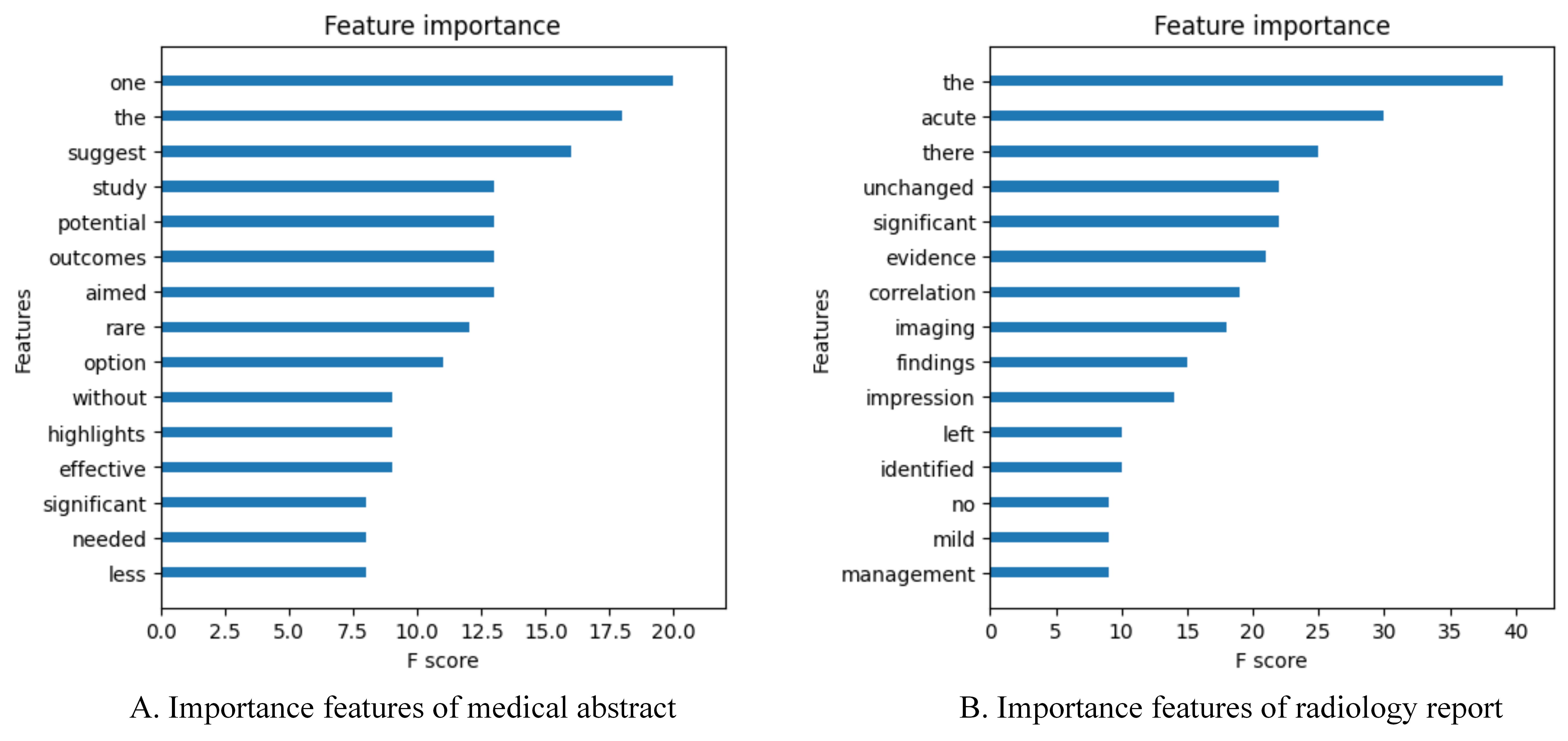} 
	\caption{Important features of the medical report and radiology report dataset. }
	\label{fig8}
\end{figure*}

\begin{figure*}[h]
	\centering
	\includegraphics[width=0.9\textwidth]{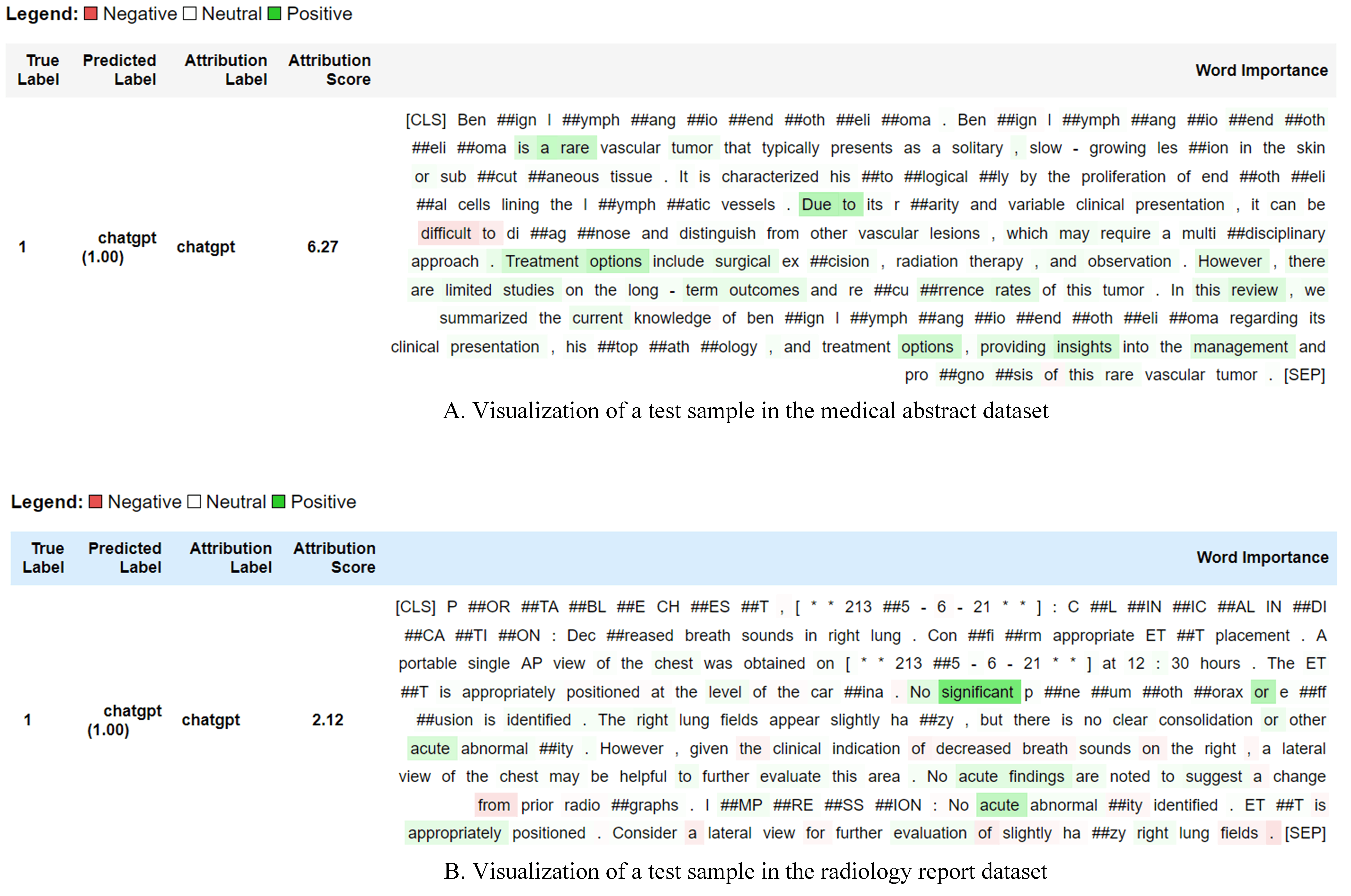} 
	\caption{Visualization of BERT. }
	\label{fig9}
\end{figure*}

Figure \ref{fig7} is the visualization of the CART model of the two data sets. It can be seen that through the decision tree with depth 4, the text generated by ChatGPT can be detected well. We define the feature importance of XGBoost as the  times of the feature appears in the  node of the base learner, and the top-20 important features are shown in Figure \ref{fig8}. Comparing Figure \ref{fig7} and Figure \ref{fig8}, we can see that their decision tree nodes are similar. For example, in the medical abstract data set, "outcomes", "study", "potential", "suggest", etc. are used as nodes in the CART and XGBoost models.

In addition to visualizing the global features of CART and XGBoost, we also use the toolkit of transformers-interpret \footnote{https://github.com/cdpierse/transformers-interpret} to visualize the local features of the samples, and the results are shown in Figure \ref{fig9}.  It can be seen that for BERT, conjuncts are  important features for detecting ChatGPT, for example: "due to", "however", "or", etc. In addition, the important features of BERT are similar to CART and XGboost. For example, "significant" and "acute" in the radiology report dataset are important features for detecting medical text generated by ChatGPT.

\section{Discussion}

\subsection{Principal Results}
In this paper, we focus on analyzing the differences between medical texts written by humans and generated by ChatGPT and design machine learning algorithms to detect medical texts generated by ChatGPT. The results show that medical texts generated by ChatGPT are more fluent and logical but more general in content and language style, while medical texts written by humans are more diverse and specific. Although ChatGPT can generate human-like text, due to the differences in their language style and content, the text written by ChatGPT can still be accurately detected by designing machine learning algorithms,  and the F1 exceeds 95\% .

\subsection{Limitations}

We only use ChatGPT as an example to analyze the difference between medical texts generated by LLM and medical texts written by humans. However, more advanced LLMs have emerged. It will be part of our future work to analyze more language styles generated by LLM and summarize the language construction rules of LLM.

\subsection{Conclusions}
In general, for AI to realize its full potential in medicine, we should not rush into its implementation, but advocate its careful introduction and open debate about risks and benefits. The medical field is a field related to human health and life. We provide a simple demonstration to identify ChatGPT-generated medical content, which can help reduce the harm caused to humans by ChatGPT-generated erroneous and incomplete information. Assessing and mitigating the risks associated with LLM and its potential harm is a complex and interdisciplinary challenge that requires combining knowledge from various fields to avoid its risks and drives the healthy development of LLM.

\bibliography{LLM_refs}

\begin{thebibliography}{10}

\bibitem{radford2018improving}
Alec Radford, Karthik Narasimhan, Tim Salimans, Ilya Sutskever, et~al.
\newblock Improving language understanding by generative pre-training.
\newblock 2018.

\bibitem{kenton2019bert}
Jacob Devlin Ming-Wei~Chang Kenton and Lee~Kristina Toutanova.
\newblock Bert: Pre-training of deep bidirectional transformers for language
  understanding.
\newblock In {\em Proceedings of NAACL-HLT}, pages 4171--4186, 2019.

\bibitem{vaswani2017attention}
Ashish Vaswani, Noam Shazeer, Niki Parmar, Jakob Uszkoreit, Llion Jones,
  Aidan~N Gomez, {\L}ukasz Kaiser, and Illia Polosukhin.
\newblock Attention is all you need.
\newblock {\em Advances in neural information processing systems}, 30, 2017.

\bibitem{brown2020language}
Tom Brown, Benjamin Mann, Nick Ryder, Melanie Subbiah, Jared~D Kaplan, Prafulla
  Dhariwal, Arvind Neelakantan, Pranav Shyam, Girish Sastry, Amanda Askell,
  et~al.
\newblock Language models are few-shot learners.
\newblock {\em Advances in neural information processing systems},
  33:1877--1901, 2020.

\bibitem{ouyang2022training}
Long Ouyang, Jeff Wu, Xu~Jiang, Diogo Almeida, Carroll~L Wainwright, Pamela
  Mishkin, Chong Zhang, Sandhini Agarwal, Katarina Slama, Alex Ray, et~al.
\newblock Training language models to follow instructions with human feedback.
\newblock {\em arXiv preprint arXiv:2203.02155}, 2022.

\bibitem{christiano2017deep}
Paul~F Christiano, Jan Leike, Tom Brown, Miljan Martic, Shane Legg, and Dario
  Amodei.
\newblock Deep reinforcement learning from human preferences.
\newblock {\em Advances in neural information processing systems}, 30, 2017.

\bibitem{schulman2017proximal}
John Schulman, Filip Wolski, Prafulla Dhariwal, Alec Radford, and Oleg Klimov.
\newblock Proximal policy optimization algorithms.
\newblock {\em arXiv preprint arXiv:1707.06347}, 2017.

\bibitem{susnjak2023applying}
Teo Susnjak.
\newblock Applying bert and chatgpt for sentiment analysis of lyme disease in
  scientific literature.
\newblock {\em arXiv preprint arXiv:2302.06474}, 2023.

\bibitem{dai2023chataug}
Haixing Dai, Zhengliang Liu, Wenxiong Liao, Xiaoke Huang, Zihao Wu, Lin Zhao,
  Wei Liu, Ninghao Liu, Sheng Li, Dajiang Zhu, et~al.
\newblock Chataug: Leveraging chatgpt for text data augmentation.
\newblock {\em arXiv preprint arXiv:2302.13007}, 2023.

\bibitem{wei2023zero}
Xiang Wei, Xingyu Cui, Ning Cheng, Xiaobin Wang, Xin Zhang, Shen Huang, Pengjun
  Xie, Jinan Xu, Yufeng Chen, Meishan Zhang, et~al.
\newblock Zero-shot information extraction via chatting with chatgpt.
\newblock {\em arXiv preprint arXiv:2302.10205}, 2023.

\bibitem{liu2023deid}
Zhengliang Liu, Xiaowei Yu, Lu~Zhang, Zihao Wu, Chao Cao, Haixing Dai, Lin
  Zhao, Wei Liu, Dinggang Shen, Quanzheng Li, et~al.
\newblock Deid-gpt: Zero-shot medical text de-identification by gpt-4.
\newblock {\em arXiv preprint arXiv:2303.11032}, 2023.

\bibitem{shen2023chatgpt}
Yiqiu Shen, Laura Heacock, Jonathan Elias, Keith~D Hentel, Beatriu Reig, George
  Shih, and Linda Moy.
\newblock Chatgpt and other large language models are double-edged swords,
  2023.

\bibitem{hisan2023chatgpt}
Urfa~Khairatun Hisan and Muhammad~Miftahul Amri.
\newblock Chatgpt and medical education: A double-edged sword.
\newblock {\em Journal of Pedagogy and Education Science}, 2(01), 2023.

\bibitem{mitrovic2023chatgpt}
Sandra Mitrovi{\'c}, Davide Andreoletti, and Omran Ayoub.
\newblock Chatgpt or human? detect and explain. explaining decisions of machine
  learning model for detecting short chatgpt-generated text.
\newblock {\em arXiv preprint arXiv:2301.13852}, 2023.

\bibitem{gilson2023does}
Aidan Gilson, Conrad~W Safranek, Thomas Huang, Vimig Socrates, Ling Chi,
  Richard~Andrew Taylor, David Chartash, et~al.
\newblock How does chatgpt perform on the united states medical licensing
  examination? the implications of large language models for medical education
  and knowledge assessment.
\newblock {\em JMIR Medical Education}, 9(1):e45312, 2023.

\bibitem{holmes2023evaluating}
Jason Holmes, Zhengliang Liu, Lian Zhang, Yuzhen Ding, Terence~T Sio, Lisa~A
  McGee, Jonathan~B Ashman, Xiang Li, Tianming Liu, Jiajian Shen, et~al.
\newblock Evaluating large language models on a highly-specialized topic,
  radiation oncology physics.
\newblock {\em arXiv preprint arXiv:2304.01938}, 2023.

\bibitem{bickmore2018patient}
Timothy~W Bickmore, Ha~Trinh, Stefan Olafsson, Teresa~K O'Leary, Reza Asadi,
  Nathaniel~M Rickles, and Ricardo Cruz.
\newblock Patient and consumer safety risks when using conversational
  assistants for medical information: an observational study of siri, alexa,
  and google assistant.
\newblock {\em Journal of medical Internet research}, 20(9):e11510, 2018.

\bibitem{liu2019roberta}
Yinhan Liu, Myle Ott, Naman Goyal, Jingfei Du, Mandar Joshi, Danqi Chen, Omer
  Levy, Mike Lewis, Luke Zettlemoyer, and Veselin Stoyanov.
\newblock Roberta: A robustly optimized bert pretraining approach.
\newblock {\em arXiv preprint arXiv:1907.11692}, 2019.

\bibitem{lan2019albert}
Zhenzhong Lan, Mingda Chen, Sebastian Goodman, Kevin Gimpel, Piyush Sharma, and
  Radu Soricut.
\newblock Albert: A lite bert for self-supervised learning of language
  representations.
\newblock {\em arXiv preprint arXiv:1909.11942}, 2019.

\bibitem{radford2019language}
Alec Radford, Jeffrey Wu, Rewon Child, David Luan, Dario Amodei, Ilya
  Sutskever, et~al.
\newblock Language models are unsupervised multitask learners.
\newblock {\em OpenAI blog}, 1(8):9, 2019.

\bibitem{lewis2020bart}
Mike Lewis, Yinhan Liu, Naman Goyal, Marjan Ghazvininejad, Abdelrahman Mohamed,
  Omer Levy, Veselin Stoyanov, and Luke Zettlemoyer.
\newblock Bart: Denoising sequence-to-sequence pre-training for natural
  language generation, translation, and comprehension.
\newblock In {\em Proceedings of the 58th Annual Meeting of the Association for
  Computational Linguistics}, pages 7871--7880, 2020.

\bibitem{raffel2020exploring}
Colin Raffel, Noam Shazeer, Adam Roberts, Katherine Lee, Sharan Narang, Michael
  Matena, Yanqi Zhou, Wei Li, and Peter~J Liu.
\newblock Exploring the limits of transfer learning with a unified text-to-text
  transformer.
\newblock {\em The Journal of Machine Learning Research}, 21(1):5485--5551,
  2020.

\bibitem{liao2023mask}
Wenxiong Liao, Zhengliang Liu, Haixing Dai, Zihao Wu, Yiyang Zhang, Xiaoke
  Huang, Yuzhong Chen, Xi~Jiang, Dajiang Zhu, Tianming Liu, et~al.
\newblock Mask-guided bert for few shot text classification.
\newblock {\em arXiv preprint arXiv:2302.10447}, 2023.

\bibitem{cai2022coarse}
Homgmin Cai, Wenxiong Liao, Zhengliang Liu, Xiaoke Huang, Yiyang Zhang, Siqi
  Ding, Sheng Li, Quanzheng Li, Tianming Liu, and Xiang Li.
\newblock Coarse-to-fine knowledge graph domain adaptation based on
  distantly-supervised iterative training.
\newblock {\em arXiv preprint arXiv:2211.02849}, 2022.

\bibitem{liu2022survey}
Zhengliang Liu, Mengshen He, Zuowei Jiang, Zihao Wu, Haixing Dai, Lian Zhang,
  Siyi Luo, Tianle Han, Xiang Li, Xi~Jiang, et~al.
\newblock Survey on natural language processing in medical image analysis.
\newblock {\em Zhong nan da xue xue bao. Yi xue ban= Journal of Central South
  University. Medical Sciences}, 47(8):981--993, 2022.

\bibitem{liu2023summary}
Yiheng Liu, Tianle Han, Siyuan Ma, Jiayue Zhang, Yuanyuan Yang, Jiaming Tian,
  Hao He, Antong Li, Mengshen He, Zhengliang Liu, et~al.
\newblock Summary of chatgpt/gpt-4 research and perspective towards the future
  of large language models.
\newblock {\em arXiv preprint arXiv:2304.01852}, 2023.

\bibitem{zhao2023brain}
Lin Zhao, Lu~Zhang, Zihao Wu, Yuzhong Chen, Haixing Dai, Xiaowei Yu, Zhengliang
  Liu, Tuo Zhang, Xintao Hu, Xi~Jiang, et~al.
\newblock When brain-inspired ai meets agi.
\newblock {\em arXiv preprint arXiv:2303.15935}, 2023.

\bibitem{alsentzer2019publicly}
Emily Alsentzer, John~R Murphy, Willie Boag, Wei-Hung Weng, Di~Jin, Tristan
  Naumann, and Matthew McDermott.
\newblock Publicly available clinical bert embeddings.
\newblock {\em arXiv preprint arXiv:1904.03323}, 2019.

\bibitem{johnson2016mimic}
Alistair~EW Johnson, Tom~J Pollard, Lu~Shen, Li-wei~H Lehman, Mengling Feng,
  Mohammad Ghassemi, Benjamin Moody, Peter Szolovits, Leo Anthony~Celi, and
  Roger~G Mark.
\newblock Mimic-iii, a freely accessible critical care database.
\newblock {\em Scientific data}, 3(1):1--9, 2016.

\bibitem{lee2020biobert}
Jinhyuk Lee, Wonjin Yoon, Sungdong Kim, Donghyeon Kim, Sunkyu Kim, Chan~Ho So,
  and Jaewoo Kang.
\newblock Biobert: a pre-trained biomedical language representation model for
  biomedical text mining.
\newblock {\em Bioinformatics}, 36(4):1234--1240, 2020.

\bibitem{luo2022biogpt}
Renqian Luo, Liai Sun, Yingce Xia, Tao Qin, Sheng Zhang, Hoifung Poon, and
  Tie-Yan Liu.
\newblock Biogpt: generative pre-trained transformer for biomedical text
  generation and mining.
\newblock {\em Briefings in Bioinformatics}, 23(6), 2022.

\bibitem{rezayi2022agribert}
Saed Rezayi, Zhengliang Liu, Zihao Wu, Chandra Dhakal, Bao Ge, Chen Zhen,
  Tianming Liu, and Sheng Li.
\newblock Agribert: knowledge-infused agricultural language models for matching
  food and nutrition.
\newblock IJCAI, 2022.

\bibitem{rezayi2022clinicalradiobert}
Saed Rezayi, Haixing Dai, Zhengliang Liu, Zihao Wu, Akarsh Hebbar, Andrew~H
  Burns, Lin Zhao, Dajiang Zhu, Quanzheng Li, Wei Liu, et~al.
\newblock Clinicalradiobert: Knowledge-infused few shot learning for clinical
  notes named entity recognition.
\newblock In {\em Machine Learning in Medical Imaging: 13th International
  Workshop, MLMI 2022, Held in Conjunction with MICCAI 2022, Singapore,
  September 18, 2022, Proceedings}, pages 269--278. Springer, 2022.

\bibitem{liu2023context}
Zhengliang Liu, Xinyu He, Lei Liu, Tianming Liu, and Xiaoming Zhai.
\newblock Context matters: A strategy to pre-train language model for science
  education.
\newblock {\em arXiv preprint arXiv:2301.12031}, 2023.

\bibitem{sezgin2022operationalizing}
Emre Sezgin, Joseph Sirrianni, and Simon~L Linwood.
\newblock Operationalizing and implementing pretrained, large artificial
  intelligence linguistic models in the us health care system: outlook of
  generative pretrained transformer 3 (gpt-3) as a service model.
\newblock {\em JMIR medical informatics}, 10(2):e32875, 2022.

\bibitem{homolak2023opportunities}
Jan Homolak.
\newblock Opportunities and risks of chatgpt in medicine, science, and academic
  publishing: a modern promethean dilemma.
\newblock {\em Croatian Medical Journal}, 64(1):1--3, 2023.

\bibitem{biswas2023chatgpt}
Som Biswas.
\newblock Chatgpt and the future of medical writing, 2023.

\bibitem{patel2023chatgpt}
Sajan~B Patel, Kyle Lam, and Michael Liebrenz.
\newblock Chatgpt: Friend or foe.
\newblock {\em Lancet Digit. Health}, 5:e102, 2023.

\bibitem{carlini2021extracting}
Nicholas Carlini, Florian Tramer, Eric Wallace, Matthew Jagielski, Ariel
  Herbert-Voss, Katherine Lee, Adam Roberts, Tom~B Brown, Dawn Song, Ulfar
  Erlingsson, et~al.
\newblock Extracting training data from large language models.
\newblock In {\em USENIX Security Symposium}, volume~6, 2021.

\bibitem{guo2023close}
Biyang Guo, Xin Zhang, Ziyuan Wang, Minqi Jiang, Jinran Nie, Yuxuan Ding,
  Jianwei Yue, and Yupeng Wu.
\newblock How close is chatgpt to human experts? comparison corpus, evaluation,
  and detection.
\newblock {\em arXiv preprint arXiv:2301.07597}, 2023.

\bibitem{bird2009natural}
Steven Bird, Ewan Klein, and Edward Loper.
\newblock {\em Natural language processing with Python: analyzing text with the
  natural language toolkit}.
\newblock " O'Reilly Media, Inc.", 2009.

\bibitem{chen2016xgboost}
Tianqi Chen and Carlos Guestrin.
\newblock Xgboost: A scalable tree boosting system.
\newblock In {\em Proceedings of the 22nd acm sigkdd international conference
  on knowledge discovery and data mining}, pages 785--794, 2016.

\bibitem{nagy2012words}
William Nagy and Dianna Townsend.
\newblock Words as tools: Learning academic vocabulary as language acquisition.
\newblock {\em Reading research quarterly}, 47(1):91--108, 2012.

\end{thebibliography}
\bibliographystyle{unsrt}

\end{document}